\documentclass[runningheads,a4paper]{llncs}

\usepackage{amssymb}
\usepackage{amsmath}
\usepackage{graphicx}
\usepackage{subcaption}

\usepackage{url}
\urldef{\mailsa}\path|{rghosh92,|
\urldef{\mailsb}\path| tangsy935,mahdi.rasouli,sinapsedirector,sunilkukreja.sinapse}@gmail.com|

\usepackage{url}
\urldef{\mailsa}\path|{rghosh92,tangsy935,mahdi.rasouli,|
\urldef{\mailsb}\path| sinapsedirector,sunilkukreja.sinapse}@gmail.com|
\newcommand{\keywords}[1]{\par\addvspace\baselineskip
\noindent\keywordname\enspace\ignorespaces#1}

\begin{document}

\mainmatter  

\title{Pose-invariant object recognition for event-based vision with slow-ELM}

\titlerunning{Pose-invariant object recognition for event-based vision with slow-ELM}

%
%
\author{Rohan Ghosh%
\and Tang Siyi \and Mahdi Rasouli \and Nitish V. Thakor\and\\
Sunil L. Kukreja}
\authorrunning{Pose-invariant object recognition for event-based vision with slow-ELM}

\institute{Singapore Institute for Neurotechnology\\
National University of Singapore, Singapore\\
\mailsa\\
\mailsb\\
}

%
%

\toctitle{}
\maketitle


\begin{abstract}
Neuromorphic image sensors produce activity-driven spiking output at every pixel. These low-power consuming imagers which encode visual change information in the form of spikes help reduce computational overhead and realize complex real-time systems; object recognition and pose-estimation to name a few. However, there exists a lack of algorithms in event-based vision aimed towards capturing invariance to transformations. In this work, we propose a methodology for recognizing objects invariant to their pose with the Dynamic Vision Sensor (DVS). A novel slow-ELM architecture is proposed which combines the effectiveness of Extreme Learning Machines and Slow Feature Analysis. The system can perform $10,000$ classifications per second, and achieves 1\% classification error for 8 objects with views accumulated over 90 degrees of 2D pose.  
\end{abstract}

\section{Introduction}
Conventional frame-based sensors capture intensity values of the whole pixel array at fixed time intervals. In contrast, asynchronous imagers remove the notion of a frame by essentially being responsive to intensity changes at an almost continual time-scale. As an example, the Dynamic Vision Sensor (DVS) elicits a spike event at a pixel when the pixel records a relative change in intensity. With their sparse, non-redundant input data stream only capturing salient moving edges, computational burden is reduced by only computing with the \textit{active events} at any time as in \cite{shape_haptic}. For object recognition this points to faster inference as highlighted in \cite{Ref_10}, wherein a few spikes acquired from moving objects enable the architecture to estimate object class. The high temporal resolution of $\approx 1 \mu s$ also allows for accurate pose-estimation in real-time when the underlying edge-structure of the object is known as shown in \cite{Ref_22}.

This work proposes a method for pose-invariant object recognition with event-based visual data. Like in \cite{pose_face} where separate eigen-faces were found pertaining to each pose, each object class is subdivided into multiple pose-specific classes. Here we use a variant of Extreme Learning Machines \cite{ELM_Theo} for classification. ELMs have shown a faster way of training neural networks, exhibiting universal approximation capabilities with their random projection based feedforward model. Our approach involves an ELM architecture with excess hidden random projections. Since not all random projections are useful for classification, we proceed to add a layer that separates the noisy and irrelevant subspaces of the projections, stripping the feature vector to a much smaller dimensional space. Quantifying the utility of a projection is not easy, but however slow feature analysis (SFA, \cite{Ref_25,SFA_rich}) proposes a simple way of arriving at informative and invariant features. For frame-based vision, SFA has been successfully applied before to learn pose-invariant features in \cite{sfa_object}.  The slowness principle targets only smoothly changing features with time, and can therefore be used to derive feature spaces which are robust to transformations. By recording data linearly varying over 2D-pose, we are able to apply the slowness principle in arriving at robust, time-supervised features. Furthermore, our constant event number sampling of events introduced in \cite{Ref_13} allows a consistent object representation which enhances recognition performance.The slow-ELM architecture proposed therefore learns to identify robust features from the recorded data exhibiting gradual 2D pose transformations of objects.

As the DVS only responds to changes, one can only expect spikes generated by the object edges when either the object or the camera is in motion. Thereby, the invariance of our classifier performance to speed is demonstrated, along with quantifying the amount of multi-pose-view information needed to make reliable class estimates. Our Slow-ELM learner shows a considerable improvement in classification performance compared to the standard ELM, achieving 1\% error with 8 objects, with their 2D pose views spanning 90 degrees. Compared to the principal components based projections as used in P-ELM \cite{pelm}, slow projections are found to give better recognition estimates. Furthermore, the system is capable of classifying $10^4 $ times per second, allowing real-time operation. For frame-based vision such high speeds are of not much use due to the 30 FPS input itself, unless there are other computational modules involved which benefit from fast classification. However, for event based vision the high temporal resolution essentially means a frame rate of $\approx 15,000$, which emphasizes the importance of fast computational modules. 


\begin{figure}[h!]
    \centering
   \includegraphics[width=0.5\textwidth]{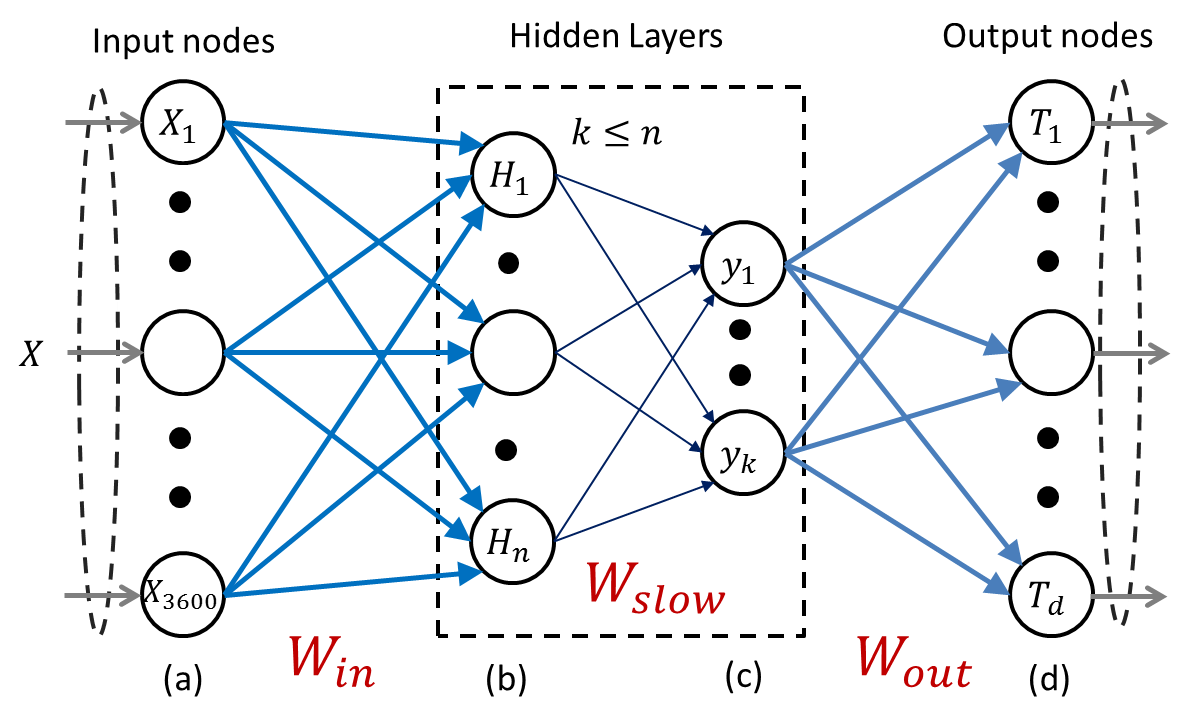}
   \caption{The Slow-ELM architecture. The transformation represented by the matrix $W_{slow}$ only preserves the slowly changing projections of $H$. $W_{in}$ correspond to the Gaussian randomized weights as in conventional ELM. $W_{out}$ is learnt between the projected signal and the output vectors. }
\label{fig:slowelm}
\end{figure}


\section{Methods}
The algorithm consists of four steps: Spatiotemporal ROI estimation; slow-ELM; pose-specific labelling; multi-view object class estimation.

\subsection{Spatiotemporal ROI Estimation}
This consists of estimating both the temporal and the spatial ROI. To obtain temporal ROI we employ the constant event number approach used in our previous work in \cite{Ref_13} which maintains event structure w.r.t change of speed. Similar to \cite{Ref_13}, a rectangular spatial ROI is obtained by considering a certain fraction of the events on each side (up, down, left, and right) of the centroid of the extracted events. Once the current spatio-temporal ROI events have been obtained, we disregard the temporal differences between those events and form a purely spatial binary image. Differently to \cite{Ref_13}, however, we add a smoothness prior to the way the ROIs change through time. This involves only including the events which are lesser than a threshold distance to the previous spatial ROI's edges. The image formed by the pixels within the ROI was then resized to a square image of a fixed size before passing onto the Slow-ELM.

\subsection{Slow-ELM}
We have training samples $\{(x_i,t_i)\}_{i=1}^{N}$ , where $(x_i)_{i=1}^{N}$ are the binary images obtained from the ROIs. $t_i$ is the target object class vector assigned to $x_i$. Every dimension of $x_i$ is scaled to the range [-1,1] before passing onto the ELM. The ELM is initialized with the entries of the input layer weights in being initialized randomnly according to the normal distribution $N(0,1)$. The $n$ hidden neuron values in $H_i$ are computed via adding a sigmoidal non-linearity $f$ onto the random projections as follows
\begin{equation}\label{eq:proj_layer1}
H_i=f(W_{in}^{T} x_i)
\end{equation}
Now the SFA algorithm elaborated in \cite{Ref_25} is applied, which finds uncorrelated linear projections of as expressed by the projection matrix :
\begin{equation}\label{eq:proj_selm}
Y_i=W_{slow}^{T}H_i
\end{equation}
The elements of $W_{slow}$ are found according to the SFA optimization method. In particular SFA looks for projections which minimize: 
\begin{equation}\label{eq:sfa_min}
\langle (\Delta y_j)^2 \rangle
\end{equation}
 Under the constraints:
 \begin{equation} \label{eq:sfa_contraints}
 \langle y_j \rangle =0   
 \end{equation}
\begin{equation}
 \langle y_{j}^2 \rangle = 1
\end{equation}
\begin{equation}
 \langle y_i y_j \rangle = 0,  i \neq j
\end{equation} 
$\langle y \rangle $ denotes the expectation of $y$ over time, in our case being the average value of the projection across all classes. The unit variance condition ensures projections stay informative. $\langle (\Delta y_j)^2 \rangle$ is the squared energy of the difference of a projection over two consecutive instances of input (difference energy). In our experiments, two consecutive instances of input only differ in the 2D-pose of the object. As noted in \cite{Ref_25}, these slow features can be obtained simply by sphering the data followed by finding the lowest eigenvalues of the difference data $\Delta y$.  As the hidden neuron vector $H$ is n-dimensional, $W_{slow}$ will be an ($n$x$n$) matrix with each column being a projection found through SFA. Since SFA returns the projections in order of decreasing difference energies we only keep the first $k$ columns of $W_{slow}$. 

\subsection{Pose-specific Labelling}
Every object data captured is categorized differently according to the 2D pose range it belongs in as we record from all viewpoints across 360 degrees (Fig.~\ref{fig:setup}a). In particular, we take 8 uniform partitions of the 2D pose: (0$^{\circ}$-45$^{\circ}$), (45$^{\circ}$-90$^{\circ}$), .. (315$^{\circ}$-360$^{\circ}$). So with N objects, we have 8N classes. The algorithm up to this point remains unsupervised as the only learning happens for finding the entries of $W_{slow}$. As shown in Fig.~\ref{fig:slowelm} the final layer is learnt through the regularized least squares algorithm shown in \cite{ELM_regclass}. For each training sample $x_i$, we extract the slow projections $Y_i$ through the aforementioned steps. Now the supervised RLS algorithm estimates the linear mapping between $Y_i$ and $t_i$ , in $W_{out}$ as used in \cite{ELM_Theo}:
\begin{equation}\label{eq:RLS}
W_{out}=(\dfrac{I}{C} + Y^TY)^{-1} Y^T T 
\end{equation}
Here $Y=[Y_1,Y_2,...,Y_N]$ and $T=[t_1,t_2,...,t_N]^T$. The parameter $C$ controls the tradeoff between the regularization and the error term. Higher the value of $C$, lesser the smoothness constraint on the weights and therefore higher the chance of over-fitting the data. Given the input to the final layer $Y_i$ we then finally end up with the output vector $t_i$:

\begin{equation}\label{eq:output}
t_i=W_{out}^{T}Y_i
\end{equation}
The class estimate is then the object for which one of its pose-specific class has the maximum value across all 8N classes in $t$.

\subsection{Multi-view object class estimation}
\label{sec:multi-view}

This describes the method used to estimate object class when multiple input data $(X_1,X_2,...,X_N)$ derived from many view-points of a single object is presented to the classifier. Since we record the event data with the object smoothly changing in pose, $(X_1,X_2,...,X_N)$ are the successive instances of the event-structure as the object rotates. The estimated object class is the one receiving the maximum number of votes across the $N$ samples, where the $i^{th}$ vote cast is to the object category inferred by the slow-ELM for $X_i$. 
\section{Experimental Setup}

\begin{figure}[h!] 
    \centering
   \includegraphics[width=1\textwidth]{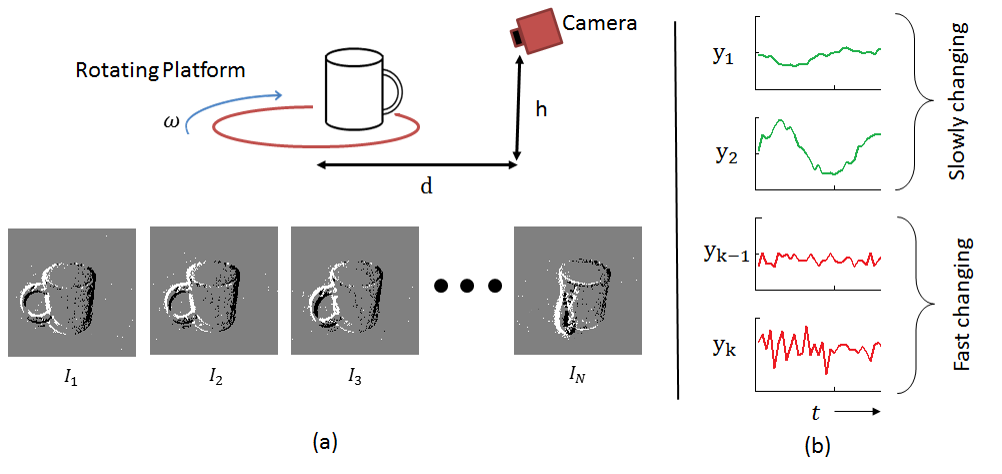}
   \caption{(a) shows the experimental setup along with sample framed event data to ($I_1$ to $I_N$) for the rotating cup object. The recording is repeated for 3 values of distance $d$ and two different heights $h$ of the camera, each time with 3 motor speeds of rotation $\omega$. (b) shows the contrast between the slowly changing and fast changing projections in response to the rotating object.}
 \label{fig:setup}
\end{figure}

As the DVS only responds to changes in the scene, the experimental setup consisted of a rotating platform on which an object was placed. Such a setup however makes the pixels near the centre of rotation generate lesser spike-events than the pixels near the edge. To avoid this motion intensity bias, the objects were placed near the edge of the platform (as shown in Fig.~\ref{fig:setup}a). For each object, the event data was captured as the platform was rotated over 6$\pi$ radians, thus uniformly covering the range of 2D-pose. The experiment was repeated for two elevations (10 cm and 40 cm) of the camera, similar to what was done in \cite{multi_view}, and across 3 different distances from the platform centre (30 cm, 45 cm, 60 cm), giving a total of 18 data recordings. For each configuration, object data was recorded for 3 different angular velocities of the platform, with a total of 8 objects. The objects chosen were: camera, cup, computer mouse, pen, mobile phone, scissors, spectacle and bottle. The output weight matrix learns a 64-class classification problem.

\section{Results and Discussion}
Out of the 18 recordings, 9 were used for testing (40 cm elevation) and the other 9 for training (10 cm elevation). Not every object had the same number of data, as they generated spikes at different event rates. Therefore for an unbiased estimate of performance, testing data for the classes having lesser examples were duplicated randomly to ensure equal instances of each class. After duplication, each class had
approximately 2700 samples. The image extracted from the ROI is resized to a 60x60 image, input as a 3600 dimensional vector to the ELM. $W_{in}$ is chosen such that $H$ has 3000 projections. We try a range of values of $k$, i.e. the dimensionality of the final vector $y$ input to the classification layer.


\begin{figure}
    \centering
   \includegraphics[width=0.5\textwidth]{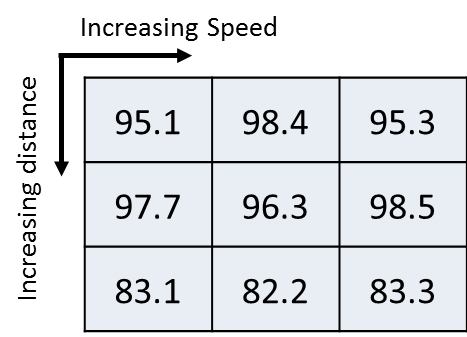}
   \caption{Recognition accuracy across 3 different speeds and distances.}
 \label{fig:mat}
\end{figure}

\subsection{Performance with varying speed and distance}
Shown in Fig.~\ref{fig:mat} is the effect of changing speeds and distance of the platform on the accuracy. The accuracy remains high for distances $d=$ 30 and 45 cm, but drops abruptly for $d=$60 cm. This indicates that the classes become less separable quickly as the distance to the object is increased beyond a limit. The effect with varying speed of the motor of the platform however is not discernible which indicates the invariance to speed changes.

\begin{figure}[htb]
\centering
  \begin{subfigure}[b]{.47\linewidth}
    \centering
    \includegraphics[width=.99\textwidth]{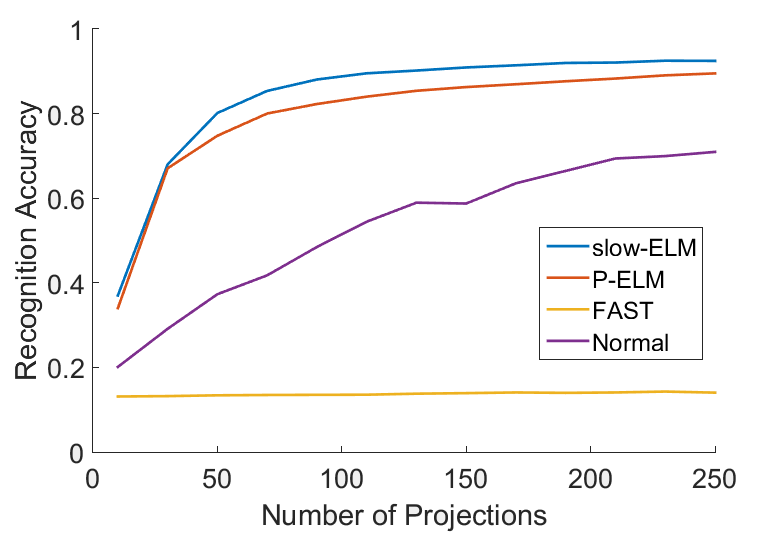}
    \caption{}\label{fig:proj}
  \end{subfigure}%
  \begin{subfigure}[b]{.48\linewidth}
    \centering
    \includegraphics[width=.99\textwidth]{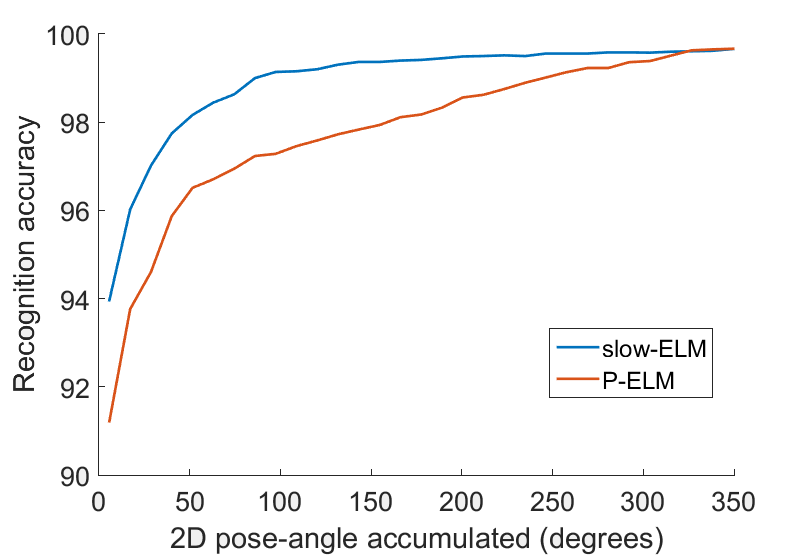}
    \caption{}\label{fig:multiview}
  \end{subfigure}%
  \caption{(a) shows Recognition Accuracy for varying number of the  $k$ selected projections used for training the output weights in $W_{out}$, shown for the different selection criteria mentioned in Section \ref{sec:criteria}b shows Recognition accuracy with slow-ELM and P-ELM on aggregated data from successive viewpoints spanning different range of 2D-pose}\label{fig:both}
\end{figure}

\subsection{Comparing slow-ELM with other selection criteria}\label{sec:criteria}
Fig.~\ref{fig:proj}. demonstrates how Slow-ELM compares in performance with traditional ELM and other variants, as a function of the number of projections used for learning. In particular, we compare slow-ELM (our approach), P-ELM \cite{pelm}, normal ELM and fast varying features (with the projections maximizing Eq.\ref{eq:sfa_min}). The figure clearly demonstrates that SFA based projections give the best recognition accuracies ( $\approx 93\%$). In contrast,
the FAST features perform very near to chance itself ( 14\%, chance is 100/64=15\%). This suggests that fast, fluctuating features do not provide abstract category information essential for classification.

\subsection{Multi-pose view object recognition}
Here the method described in Section \ref{sec:multi-view} is used to arrive at class estimates with event-data accumulated across changing pose as the objects rotate. Precisely, we quantify the recognition accuracy when event data spread out in different range of 2D pose is available. This is averaged across all possible starting 2D-poses of the objects. Fig.~\ref{fig:multiview} compares the recognition accuracy for both SFA and PCA based projections. It can be seen that SFA quickly reaches a low error rate (1\%) in classification with only 90 degrees of pose information whereas PCA requires 280 degrees to achieve the same error.

\section{Conclusion}
This work presents a system capable of recognizing objects from a real-time feed of spike-events and capable of generating accurate class estimates by combining information from successive views varying in object pose. Apart from the low computation time which allows upto $10^4$ classifications per second, the training time is also considerably lesser than the state-of-the-art Convolutional Neural Networks. The speed invariance and the partial scale invariance (object distance) of the classifier has been demonstrated. A novel slow-ELM architecture has been proposed to extract features invariant to pose changes. 
\bibliographystyle{unsrt}
\bibliography{main}



\end{document}